\documentclass[10pt,twocolumn,letterpaper]{article}

\usepackage{cvpr}
\usepackage{times}
\usepackage{epsfig}
\usepackage{graphicx}
\usepackage{amsmath}
\usepackage{amssymb}

\usepackage[breaklinks=true,bookmarks=false]{hyperref}

\cvprfinalcopy 

\def\cvprPaperID{24} 

\begin{document}

\title{Tiny-Inception-ResNet-v2: Using Deep Learning for Eliminating Bonded Labors of Brick Kilns in  South Asia}

\author{Usman Nazir, Numan Khurshid, Muhammad Ahmed Bhimra, Murtaza Taj \\
Department of Computer Science, Syed Babar Ali School of Science and Engineering\\Lahore University of Management Sciences (LUMS), Lahore, Pakistan \\
{\tt\small \{17030059, 15060051, 17030015, murtaza.taj\}@lums.edu.pk}}

\maketitle

\begin{abstract}
   This paper proposes to employ a Inception-ResNet inspired deep learning architecture called \emph{Tiny-Inception-ResNet-v2} to eliminate bonded labor by identifying brick kilns within ``Brick-Kiln-Belt'' of South Asia. The framework is developed by training a network on the satellite imagery consisting of 11 different classes of South Asian region. The dataset developed during the process includes the geo-referenced images of brick kilns, houses, roads, tennis courts, farms, sparse trees, dense trees, orchards, parking lots, parks and barren lands. The dataset\footnote{https://cvlab.lums.edu.pk/?p=1779} is made publicly available for further research. Our proposed network architecture with very fewer learning parameters outperforms all state-of-the-art architectures employed for recognition of brick kilns. Our proposed solution would enable regional monitoring and evaluation mechanisms for the Sustainable Development Goals.
\end{abstract}


\section{Introduction}
According to the Global Slavery Index of 2018 $24.9$ million people are trapped in forced labor globally~\cite{boyd2018slavery}. An estimated $12.7$ million are within ``Brick-Kiln-Belt'' of South Asia ($1,551,997~km^2$ between Afghanistan, Pakistan, India, Bangladesh and Nepal), because indebted labor is a major contributing factor to the prevalence of slavery in the Asia-Pacific constituting 55\% of all forced labor-mapping brick kilns and estimating their labor force is an essential step in eliminating slavery from the region. The UN's Sustainable Development Goal (SDGs) 8 specifically refers to forced labor and governs the world over aims for ending modern slavery by 2025~\cite{foody2019earth}. They are, however, faced with a lack of access to reliable, up-to-date and actionable data on slavery activity. Such data needs to be spatially explicit and scalable to allow governments to monitor activities and implement strategies to emancipate individuals trapped within institutions of slavery. The absence and insufficiency of data compromises evidence-based action and policy formulation, as has been highlighted by various sustainable development data-gap analysis reports. Thus, to meet this challenge, new and innovative approaches are needed.
	\begin{figure}[t]
		\begin{tabular}{ccc}
			\includegraphics[width=.26\columnwidth]{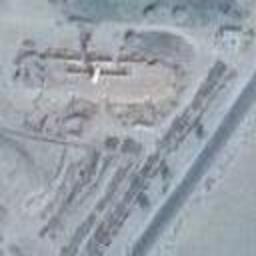} &\includegraphics[width=.26\columnwidth]{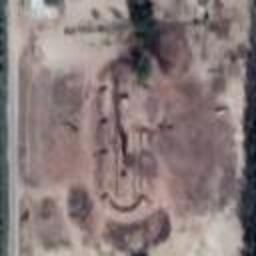}&
			\includegraphics[width=.26\columnwidth]{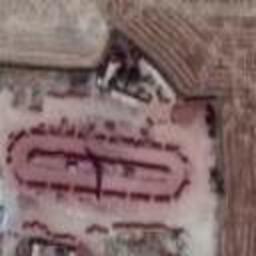}\\
			\includegraphics[width=.26\columnwidth]{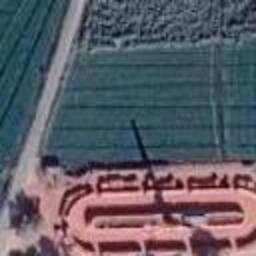} &\includegraphics[width=.26\columnwidth]{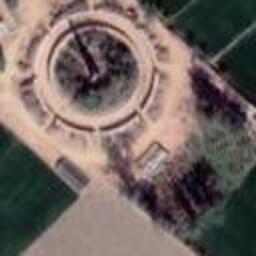}&
			\includegraphics[width=.26\columnwidth]{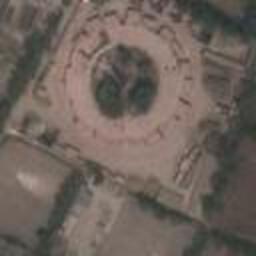}\\
			\includegraphics[width=.26\columnwidth] {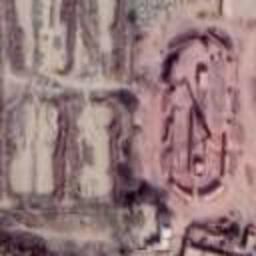} &\includegraphics[width=.26\columnwidth]{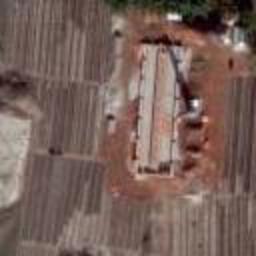}&
			\includegraphics[width=.26\columnwidth]{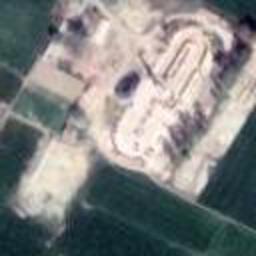}\\
		\end{tabular}
		\caption{Example satellite imagery of brick kilns from different spatial locations showing variation in quality, structure and color profile}
		\label{fig:Var}
	\end{figure}
	\begin{figure*}[h]
		\centering
		\includegraphics[scale=0.7, trim={2cm 5cm 2cm 6cm},clip]{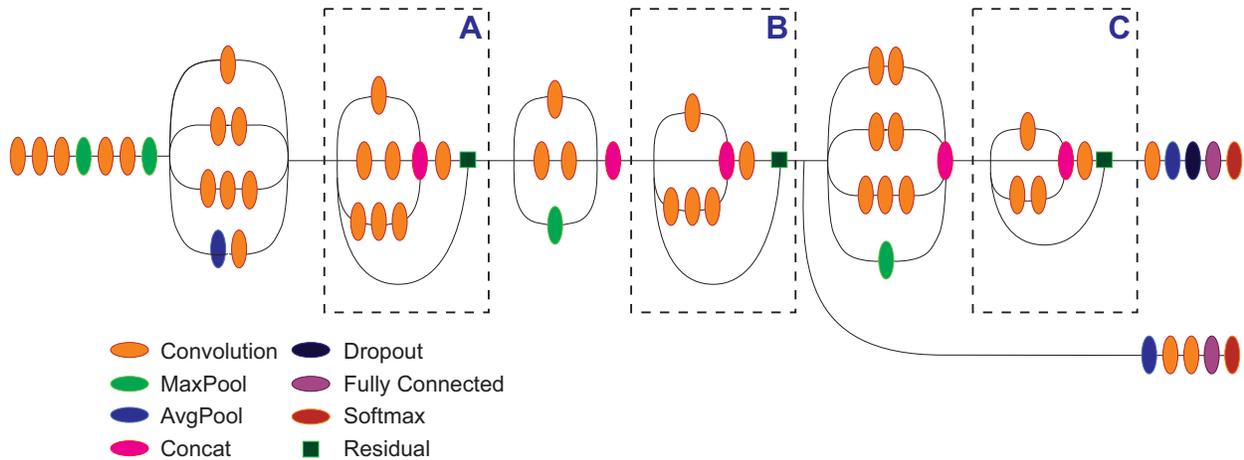}
		\caption{Inception-ResNet-V2 contains 11 A-blocks, 20 B-blocks, and 10 C-blocks and our proposed tiny version only contains 10 A-blocks, 3 B-blocks, and 3 C-blocks}
		\label{Tiny-Inception-ResNet-v2}
	\end{figure*}
	
	One of the key challenges faced by all development efforts is selecting target populations, or the beneficiaries of programs. In order to ensure that interventions reach those who need them, and have an impact on economic and human development, one must first identify those who are vulnerable and most in need of help. This problem only increases in complexity when we seek to identify individuals trapped in slavery or forced labor. In the past, household surveys have generally been the norm in identifying those in need of development assistance. However, surveys are costly and cumbersome, and rarely have large enough samples to be representative at the village or town level. Data is thus usually aggregated at the national, regional, provincial, or district level. Statistics estimated at the national or district level, however, may mask the fact that there are pockets of space which are severely in need of assistance. In addition to this, slave owners may hinder data collection in the areas that they control, or enslaved individuals may face numerous pressures in disclosing themselves as enslaved to enumerators or authorities. 
	
	Remote sensing is one such method that has been applied to various social and developing issues to help better inform policy. Remote sensing refers to the use of satellite or other sensor technologies to obtain information about the physical characteristics of an area from afar without conducting expensive and strenuous surveys. Moreover, new and improved techniques in computer vision and deep learning use large datasets with somewhat similar proxy variables as are collected in surveys to gauge vulnerability on a more minute scale and provides solution with minimal human intervention.
	
	To develop an efficient automated solution for brick kiln identification, we propose Tiny-Inception-ResNet-v2 network applied to satellite imagery.  We also developed a new satellite image dataset comprising the images of the South Asian region for 11 different categories. The dataset will be made publicly available and will help inform the policy of governments within the region, and allow them to gauge the success of past interventions, as well as serve as a baseline by which to assess future progress. To the best of our knowledge, this will be the first of its kind effort that goes beyond the geographic boundaries of a specific country and aims to provide a comprehensive survey of bonded labor at kilns along with their accurate geolocation.

	\subsection{Related Work}
	Availability of high resolution satellite imagery along with recent advancements in machine learning particularly deep convolutional neural networks (CNN) have paved a way for large scale analysis of wide variety of parameters across the globe. High resolution satellite imagery has been used by~\cite{xie2016transfer} to estimate per capita expenditure in Africa. Recently, remote sensing images have also been used to analyze the extent of modern slavery. For instance,~\cite{boyd2018slavery} have mapped fish farms suspected of using child slave labor in the Sundarbans Reserve Forest in Bangladesh. Similarly,~\cite{boyd2018slavery},\cite{jackson2018analysing}, through their ``Slavery from Space'' project proposed a crowd-sourced procedure to manually detect brick kilns from satellite imagery. However, they were only able to manually annotate 320 geographic cells (i.e. only 2\% of the entire Brick-Kiln-Belt). This study also does not account for workforce size of each kiln. 
	
	The potential inherent within remote sensing and machine learning for social science research and development purposes has been noted by many~\cite{boyd2018slavery,blumberg1997new,huo2018adversarial,lecun2015deep}. Residual Neural Network~\cite{he2016deep} have been shown to produce favorable results on satellite imagery for the problem of classifying imagery into known classes such as road, houses, vegetation etc. Similarly,~\cite{he2017mask} proposed algorithm to classify each pixel of satellite imagery into known classes thus obtaining an exact boundary of the object of interest in the given satellite image.
	
	One area that has especially benefited from such techniques is poverty studies. Machine learning has been used to identify poverty stricken populations and their geographic location through the analysis of high resolution satellite images. One such technique utilizes the premise that nighttime luminosity is strongly correlated (if noisily) with economic activity and development~\cite{jean2016combining}. Using satellite images, such efforts compare daytime and nighttime images of a specified location (a country, a continent) to identify which factors are associated with greater night lighting e.g. paved road networks.~\cite{jean2016combining} find in their study that their method of estimating consumption expenditure and asset wealth through machine learning is able to explain 75\% of the variation in local level economic outcomes. Though such methodologies cannot detect poverty at the level of the individual or household, they can inform policy makers about which sub-populations or communities are economically marginalized or vulnerable, and help them target their interventions at particular geographical locations where they are most needed. Recent contribution of \cite{foody2019earth} analyzed various machine learning techniques to automate the process of brick kiln identification in the given tile of images. However, their approach outputs very high false positive rate. To cope with the scenario they proposed to train a two-staged R-CNN classifier to achieve acceptable performance.
	
	Our approach contributes by proposing Tiny-Inception-ResNet-v2, an improved architecture inspired from Inception-ResNet-v2~\cite{szegedy2017inception}  to identify kilns in the satellite images. We showed that with fewer learnable parameters of Tiny-Inception-ResNet-v2 we could easily classify brick kilns with about $95\%$ accuracy. 

\begin{figure}[h]
	\begin{tabular}{ccc}
		\includegraphics[width=.26\columnwidth]{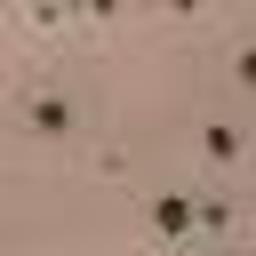} &\includegraphics[width=.26\columnwidth]{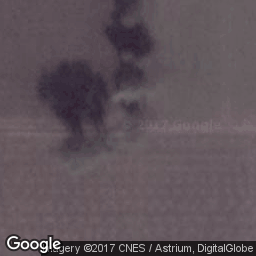}&
		\includegraphics[width=.26\columnwidth]{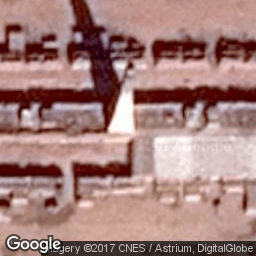}\\
		Barren Land & Sparse Trees & Brick Kiln \\
		\includegraphics[width=.26\columnwidth]{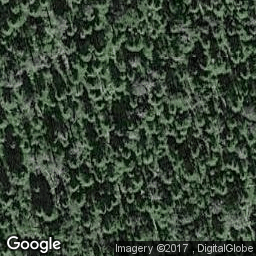} &\includegraphics[width=.26\columnwidth]{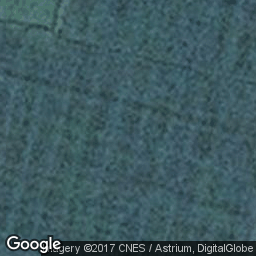}&
		\includegraphics[width=.26\columnwidth]{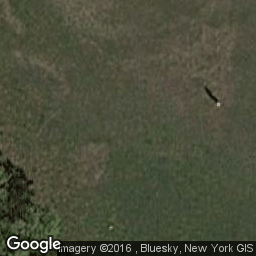}\\
		Dense Trees & Farms & Grass \\
		\includegraphics[width=.26\columnwidth] {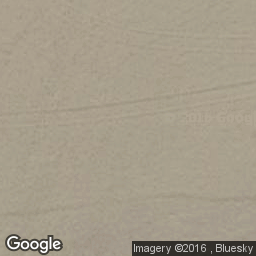} &\includegraphics[width=.26\columnwidth]{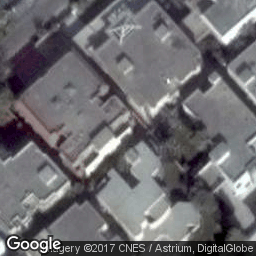}&
    	\includegraphics[width=.26\columnwidth]{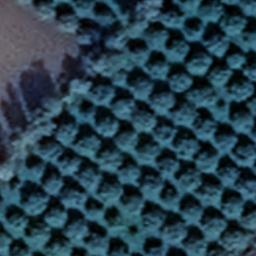}\\
    	Ground & Houses & Orchard \\
    	\includegraphics[width=.26\columnwidth] {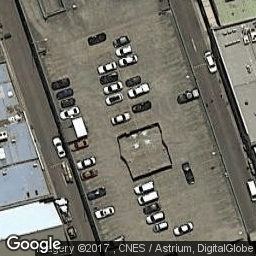} &\includegraphics[width=.26\columnwidth]{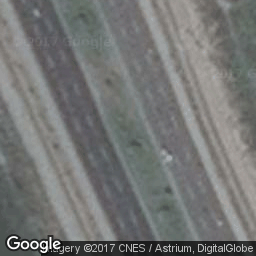}&
    	\\
    	Parking & Roads & \\
    	\end{tabular}
	\caption{Example satellite image from every class in dataset (Satellite image courtesy Google Maps)}
	\label{fig:dataset}
\end{figure}

\section{Challenges}
Identifying brick kilns from satellite imagery involves the following challenges:
\subsection{Structural Variations}
Brick kilns may significantly vary in structure, shape, and size as shown in Fig.~\ref{fig:Var}. They come in circular, oval, rectangular, and elongated structures. These variations are usually enforced by geographical locations, environmental conditions, manufacturing technologies, building material, and local building regulatory authorities. This is considered as one of the obvious challenges, for any machine learning approach, to learn a generic set of visual features for each type of kilns. 
\subsection{Environmental Variations}
Satellite images are prone to atmospheric variations including but not limited to cloud cover, pollution, variation in luminosity, and seasonal changes in the environment for which the imagery has been acquired. These variations in images contributes in confusing the classifier. An obvious effect of these changes could be observed between the top-left and top-right images of Fig.~\ref{fig:Var}. 
\subsection{Sensor Variations}
Satellite images are acquired through various imaging devices and sensors installed on satellites. Some of the famous ones include WorldView, Pleiades, GeoEye-1, and QuickBird. Sensor variations are usually caused by difference in parameters of imaging devices capturing same imagery differently over time. When spatial analysis is spread across multiple cities, these changes in sensors can be observed as huge variations in the quality, resolution and color profile of the imagery due to the different satellite data (see Fig.~\ref{fig:Var}).

\begin{figure}
    \centering
    \includegraphics[width=\columnwidth]{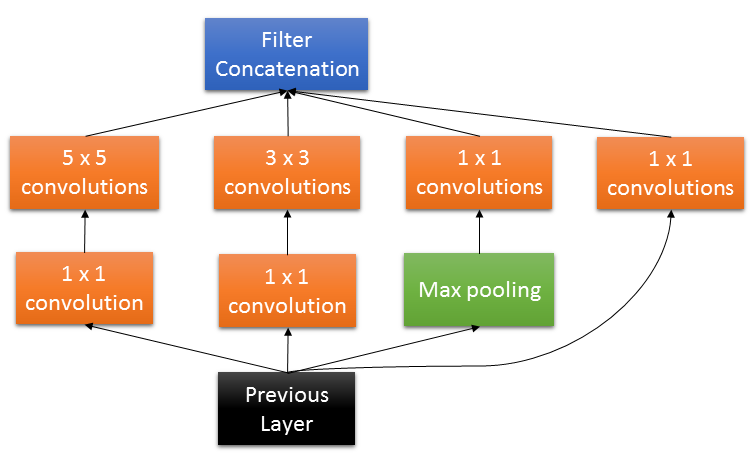}
    \caption{Inception Block}
    \label{fig:inceptionBlock}
\end{figure}

\section{Proposed Method}

\subsection{Dataset Generation and Pre-processing}
We developed a dataset of satellite imagery from South Asian region by acquiring images at zoom level 20 with their geo-locations from OpenStreetMaps (OSM). These images were then manually labelled through crowd-sourcing into 11 different classes each having around $600$ images.  
All these $6827$ images were used to train the proposed classifier for categories of kilns, houses, road, tennis courts, farms, sparse trees, dense trees, orchards, parking, parks and barren lands (see Fig.~\ref{fig:dataset}).  

To identify brick kilns in the "Brick-Kiln-Belt" we used the same network, testing it on an entirely different images from digital globe imagery ($787,100$ images) for the entire Afghanistan-to-Nepal region $1,551,997~km^2$ for the year 2018. These images acquired at zoom-level $17$ ($1.193~\frac{meters}{pixel}$ on Equator) were first converted to zoom-level $20$ (about $50$ million images) using following pre-processing steps:
\begin{enumerate}
    \item Finding the midpoint \emph{latitude-longitude} (lat-lon) of each image tile of zoom-level 17 using following equations:
        \begin{multline}
         lat^{z_{17}}=\Bigg[\Big((Y_{tile}^{z_{17}}-A)\times B\Big)
         + C\Bigg],  
        \end{multline}
        where 
        \begin{itemize}
            \item $A = 42295$ is a reference tile \#. It can be chosen randomly; however, we selected $42295$ throughout our experiments.
            \item $B = 0.0054931640625$ is difference between adjacent tiles of zoom-level 17.
            \item $C = 52.33612060546875$ is a relative lat and lon with respect to $A$.
        \end{itemize}
        \begin{multline}
         lon^{z_{17}}=\Bigg[\Big((X_{tile}^{z_{17}}-A)\times B\Big)
         +C\Bigg],   
        \end{multline}
        where $A$, $B$ and $C$ are given above.
    \item Converting zoom-level $17$ lat-lon to zoom-level 20 lat-lon using following equations:
        \begin{equation}
        cornerLat^{z_{20}}=lat^{z_{17}} +4*D,
        \end{equation}
        \begin{equation}
        cornerLon^{z_{20}}=lon^{z_{17}}-4*D,
        \end{equation}
        where $D=6.866455078125*10^{-4}$ is the difference between lat/lon of adjacent zoom-level $20$ tiles.
\end{enumerate}

In our work each image is associated with its geo-location which makes it very easy for images classified as brick kilns class to be identified on the maps. The geo-location of the identified brick kiln could then be linked to the census data provided by Punjab-Brick-Kiln-Census-dataset~\footnote{http://202.166.167.115/brick\_kiln\_dashboard/} which is publicly available containing all the relevant information regarding number of workers, number of children, age of each child and schooling status associated with each kiln. 

We proposed a classifier that is inspired from the well-known Inception-ResNet-v2 netowrk. The performance gain of the network as compare to other deep learning architectures encompassing the inception block incorporated in the network.
\subsection{Inception Block}
The  inception  module consists of convolutions of different sizes that allow the network to process features at different spatial scales. They are then lumped and fed to the next layer for further processing as shown in Fig.~\ref{fig:inceptionBlock}. For dimensionality reduction, 1x1 convolutions are used before the more expensive 3x3 and 5x5 convolutions.
In many remote-sensing problems, we need the deeper network to process features at different spatial scales. To cope with our challenges, such flexibility can be incorporated in convolutional neural networks by introducing inception blocks. 
\subsection{Tiny-Inception-ResNet-v2}
An improved form of Inception-ResNet-v2 called as "Tiny-Inception-ResNet-v2" with much less inception blocks have been proposed to classify satellite images. We suppose that even fewer number of inception blocks used in Inception-ResNet network are able to classify the satellite images in classes. For this purpose optimal number of inception blocks are to be identified. Fig.~\ref{Tiny-Inception-ResNet-v2} shows the detailed architecture of our proposed network with a block level architecture at the top of the figure while the expended version at the bottom. Three main blocks namely A, B, and C containing different number of stacked inception blocks are shown. To identify the optimal number of inception modules to be used in each block, we tried various combinations given in Table~\ref{comparison} and found that network with $10$ inception block in A, $3$ blocks in B, and $3$ blocks in C outperforms the other combinations. It has also been observed that this network has much fewer learnable parameters as compare to other combinations. More inception blocks have been used in A than B and C, to learn the features of the image responsible for visual appearance. The later ones focus on discriminating the learned features.


 \begin{table*}[ht]
\caption{Comparison between different versions of Inception-ResNet-v2}
	\centering
	\scalebox{0.9}{
	\begin{tabular}{|c|c|c|c|c|c|c|c|}
		\hline
		 A-block & B-block & C-block & Val Loss & Precison & Recall &F1 Score & Parameters  \\ \hline
	        11 & 20 & 10 & 0.00037 & \bf{0.9950}   &0.8550   & 0.9200& 54,353,643   \\
	        20 &  5 & 5 &   \bf{0.00016}   & 0.9940 & 0.2990 & 0.4597&27,243,579 \\ 
	        10 & 3   & 3 &    0.00022  &  0.9854 & \bf{0.9052} & \bf{0.9435} &19,682,011  \\ 
	        10 & 1 & 1 &  0.00041   & 0.993  & 0.3130 & 0.4760    &13,354,523\\ \hline
	\end{tabular}}
	\label{comparison}
\end{table*}

\begin{figure}[t]
    \centering
    \includegraphics[scale=0.4]{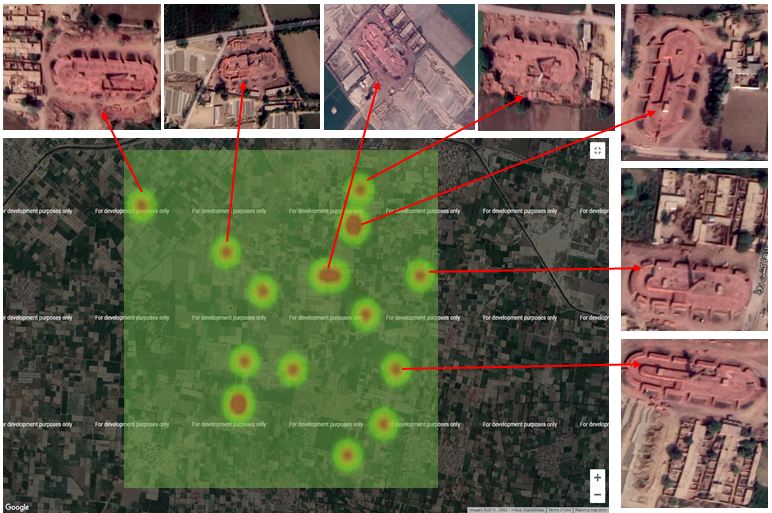}
    \caption{Qualitative Analysis on Kot Radha Kishan, in Punjab Province, Pakistan. It can be seen within the green region-of-interest; proposed network correctly classified all brick kilns with threshold 0.5. (Satellite image courtesy Google Maps)}
    \label{kotRadhaKishanQualitativeResult}
\end{figure}
\begin{figure}[t]
    \centering
    \includegraphics[scale=0.35]{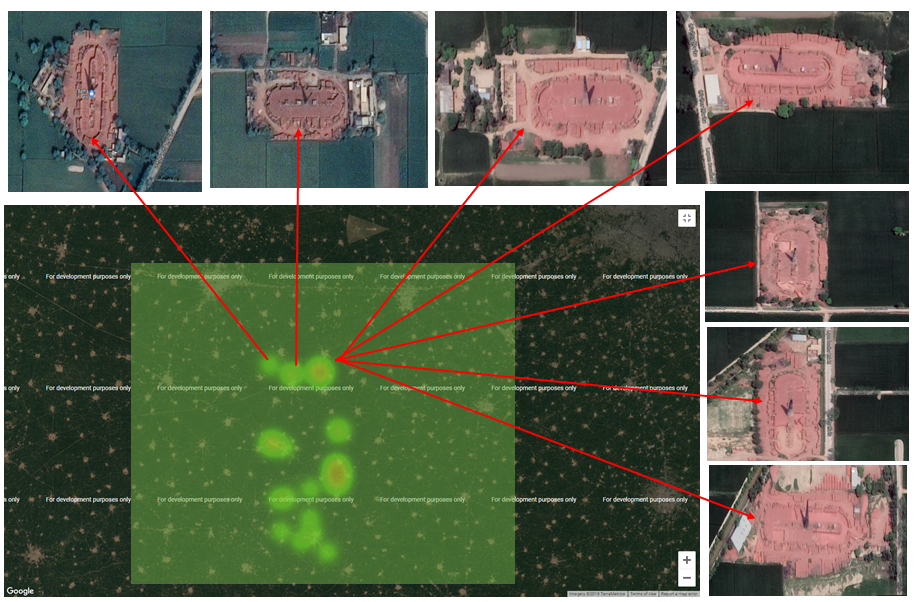}
    \caption{Qualitative Analysis on Province Punjab, India. It can be seen within the green region-of-interest; proposed network correctly classified all brick kilns with threshold 0.9. (Satellite image courtesy Google Maps)}
    \label{PunQajIndiabualitativeResult}
\end{figure}

\section{Results and Evaluation}
We evaluated our trained network of Tiny-Inception-ResNet-v2 on unseen dataset consisting of 700 images of zoom 20 for quantitative analysis and 787,100 images of zoom 17 acquired from Digital Globe Imagery for qualitative analysis. The classifier with output kiln is considered as 1 whereas the rest of the class outputs are considered as non-kiln (0). Following benchmark performance metrics have been calculated to validate the performance of our approach.
\subsection{Performance Metrics}
Precision, Recall, and F1-Score have been calculated for each of the networks as: 
    \begin{equation}
        \small{Precision = \frac{True~Positives}{True~Positives + False~Positives}}
    \end{equation}
    \begin{equation}
        \small {Recall = \frac{True~Positives}{True~Positives + False~Negatives}}
    \end{equation}
    \begin{equation}
         \small{F1~Score = \frac{2}{\frac{1}{Precision}+\frac{1}{Recall}}}
    \end{equation}

\begin{table*}[t]
	\caption{Table showing quantitative evaluation of the proposed network with state-of-the-art architectures}
	\centering
	\begin{tabular}{|c|c|c|c|c|c|c|c|}
		\hline
		\bf{Network Architectures} & \bf{Precision} &\bf{ Recall} & \bf{F1 Score} & \bf{\# of Parameters}  \\ \hline
		Two Staged R-CNN \cite{foody2019earth}&  0.9494  & 0.9494 & 0.9494 & - \\ \hline
		ResNet-152 &  0.9906  & 0.8166 & 0.8952 & 41,407,238 \\ \hline
		ResNet-50 &  \bf{0.9909}  & 0.8416 & 0.9102 & 21,312,267\\ \hline
		ResNet-34 &  0.9892  & 0.8841 & 0.9337 & 21,312,267 \\ \hline
		Inception-v3 & 0.9846 & 0.7413 & 0.8458 & 21,815,078  \\ \hline
		Inception-ResNet-v2 & 0.9955  & 0.8552 & 0.9200& 54,345,958\\ \hline
		Tiny-Inception-ResNet-v2 (Proposed) & 0.9854 & \bf{0.9052} & \bf{0.9435} & \bf{19,682,011} \\ \hline
	\end{tabular}
	\label{quantitativeEvaluation}
\end{table*}

\subsection{Quantitative Results}
The proposed method explained in the previous section to identify brick kilns has been tried for various standard deep learning residual architectures. Performance metrics with their number of learnable parameters have been shown in Table.~\ref{quantitativeEvaluation}. It is evidently clear from the table that even though deep ResNets and Inception networks may consists of many layers and much more learnable parameters, still they fail to perform for our problem. Our proposed Tiny-Inception-ResNet-v2 may lag behind in terms of precision value; however, it leads the list in recall, F1-score, and number of learnable parameters.
One of the major problems with the two staged R-CNN approach \cite{foody2019earth} was, huge number of false positives identified by the binary classifier. The issue was resolved by adapting the two-staged classification approach. Our proposed multi-class classifier has very less false positive results with a recall of about $90.52\%$ and F1-Score of $94.35\%$ with 19m parameters only.

\subsection{Qualitative Results}
For the qualitative analysis of our approach we showed the results of Kot Radha Kishan, Punjab, Pakistan in Figures~\ref{kotRadhaKishanQualitativeResult} and Punjab, India in  \ref{PunQajIndiabualitativeResult} respectively. The images clearly show that brick kilns with various structural differences have been identified by our proposed classifier. Moreover, our classifier efficiently recognizes the brick kiln in the image despite the variation in size, spatial location and orientation of the kiln in the image. The heatmap generated pin points each and every brick kiln identified in block at zoom 17 adapting the proposed method.

\section{Conclusion}
We proposed a novel network architecture: ``Tiny-Inception-ResNet-v2'' to identify the brick kilns in satellite images. The network was trained on zoom-level $20$ images  ($782$ images) of brick kilns of Lahore, Pakistan while tested on the zoom-level $17$ images of South Asia (about $700$). Despite the structural and color variation in the images of kilns, our proposed network, with very fewer learning parameters, outperforms all the state-of-the-art network architectures achieving F1-Score of around $94\%$. We also provided  detailed geo-referenced dataset and annotations for 11 classes, which will serve as a valuable resource for further such analysis.

\bibliographystyle{IEEEtran}
\bibliography{egbib}

\end{document}